\documentclass[sigconf]{acmart}

\copyrightyear{2021}
\acmYear{2021}
\setcopyright{acmcopyright}
\acmConference[MM '21] {Proceedings of the 29th ACM Int'l Conference on Multimedia}{October 20--24, 2021}{Virtual Event, China.}
\acmBooktitle{Proceedings of the 29th ACM Int'l Conference on Multimedia (MM '21), Oct. 20--24, 2021, Virtual Event, China}
\acmPrice{15.00}
\acmISBN{978-1-4503-8651-7/21/10}
\acmDOI{10.1145/3474085.3475355}

\usepackage{booktabs}
\usepackage{amsmath}
\usepackage{subfigure}
\makeatletter
\renewcommand*\env@matrix[1][\arraystretch]{%
	\edef\arraystretch{#1}%
	\hskip -\arraycolsep
	\let\@ifnextchar\new@ifnextchar
	\array{*\c@MaxMatrixCols c}}
\makeatother

\begin{document}
	\fancyhead{}

\title{Multi-initialization Optimization Network for Accurate 3D Human Pose and Shape Estimation}


\author{Zhiwei Liu$^{1,2,5*}$, Xiangyu Zhu$^{1,2*}$,  Lu Yang$^{3}$,  Xiang Yan$^{6}$,  Ming Tang$^{1,2}$, Zhen Lei$^{1,2,4}$,  Guibo Zhu$^{1,2}$,  Xuetao Feng$^{7}$,  Yan Wang$^{7}$, and Jinqiao Wang$^{1,2,5}$}
\thanks{$^\ast$Equal contribution.}
\affiliation{%
	\institution{$^1$National Laboratory of Pattern Recognition, Institute of Automation, Chinese Academy of Sciences}
	\institution{$^2$School of Artificial Intelligence, University of Chinese Academy of Sciences}
	\institution{$^3$Beijing University of Posts and Telecommunications}
	\institution{$^4$Centre for Artificial Intelligence and Robotics, Hong Kong Institute of Science and Innovation, Chinese Academy of Sciences}
	\institution{$^5$ObjectEye, Inc.}
	\institution{$^6$Dilusense Technology Corporation.}
	\institution{$^7$Alibaba Group}
}
\email{{zhiwei.liu,xiangyu.zhu}@nlpr.ia.ac.cn, soeaver@bupt.edu.cn, xiang9292@foxmail.com, {tangm,zlei,gbzhu}@nlpr.ia.ac.cn, 
{xuetao.fxt,wy84378}@alibaba-inc.com, jqwang@nlpr.ia.ac.cn}

\renewcommand{\shortauthors}{Trovato and Tobin, et al.}

\begin{abstract}
	3D human pose and shape recovery from a monocular RGB image is a challenging task. Existing learning based 
	methods highly depend on weak supervision signals, e.g. 2D and 3D joint location, 
	due to the lack of in-the-wild paired 3D supervision. 
	However, considering the 2D-to-3D ambiguities existed 
	in these weak supervision labels, the network is easy to get stuck in local optima when trained with such labels. 
	In this paper, we reduce the ambituity by optimizing multiple initializations. 
	Specifically, we propose a three-stage framework named Multi-Initialization Optimization Network (MION). 
	In the first stage, we strategically select different coarse 3D reconstruction 
	candidates which are compatible with the 2D keypoints of input sample. 
	Each coarse reconstruction can be regarded as an initialization leads to one optimization branch. 
	In the second stage, we design a mesh refinement transformer (MRT) to 
	respectively refine each coarse reconstruction result via a self-attention mechanism. Finally, a Consistency Estimation 
	Network (CEN) is proposed to find the best result from mutiple candidates by evaluating 
	if the visual evidence in RGB image matches a given 3D reconstruction. 
	Experiments demonstrate that our Multi-Initialization Optimization Network 
	outperforms existing 3D mesh based methods on multiple public benchmarks.

\end{abstract}

\ccsdesc[300]{Human-centered computing}
\ccsdesc[500]{Computing methodologies~Shape inference}
\ccsdesc[500]{Computing methodologies~Reconstruction}

\keywords{3D human reconstruction, 3D pose estimation, deep learning }


\maketitle

\begin{figure}
	\centering
	\subfigure[2D-to-3D ambiguity: different 3D results which are matched with the same 2D keypoints.]{
		\begin{minipage}[b]{0.45\textwidth}
			\includegraphics[width=\linewidth]{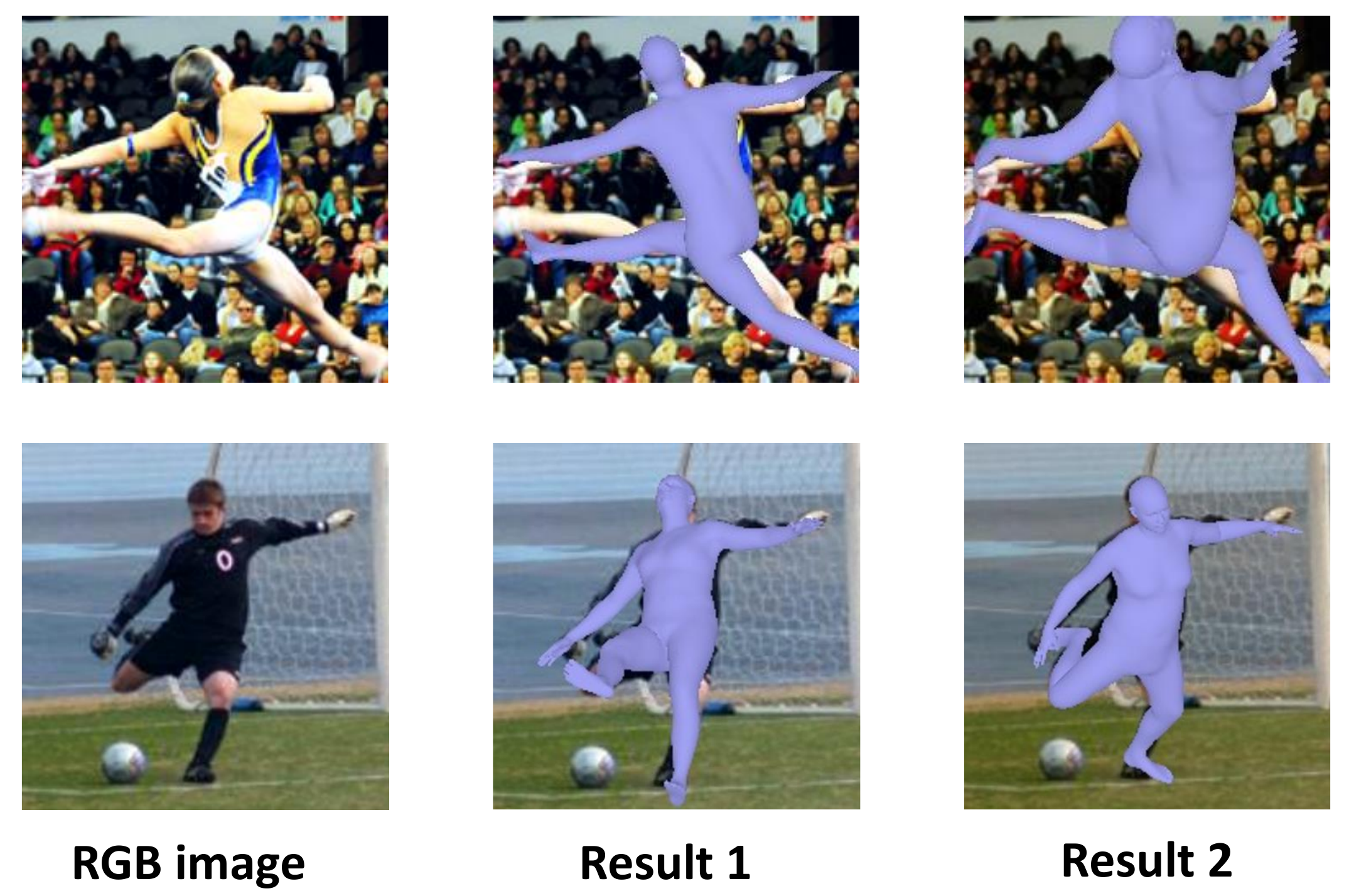} \\
		\end{minipage}
	}
	\subfigure[Comparsion between traditional regression based method and our Multi-initialization Optimization Network]{
		\begin{minipage}[b]{0.45\textwidth}
			\includegraphics[width=\linewidth]{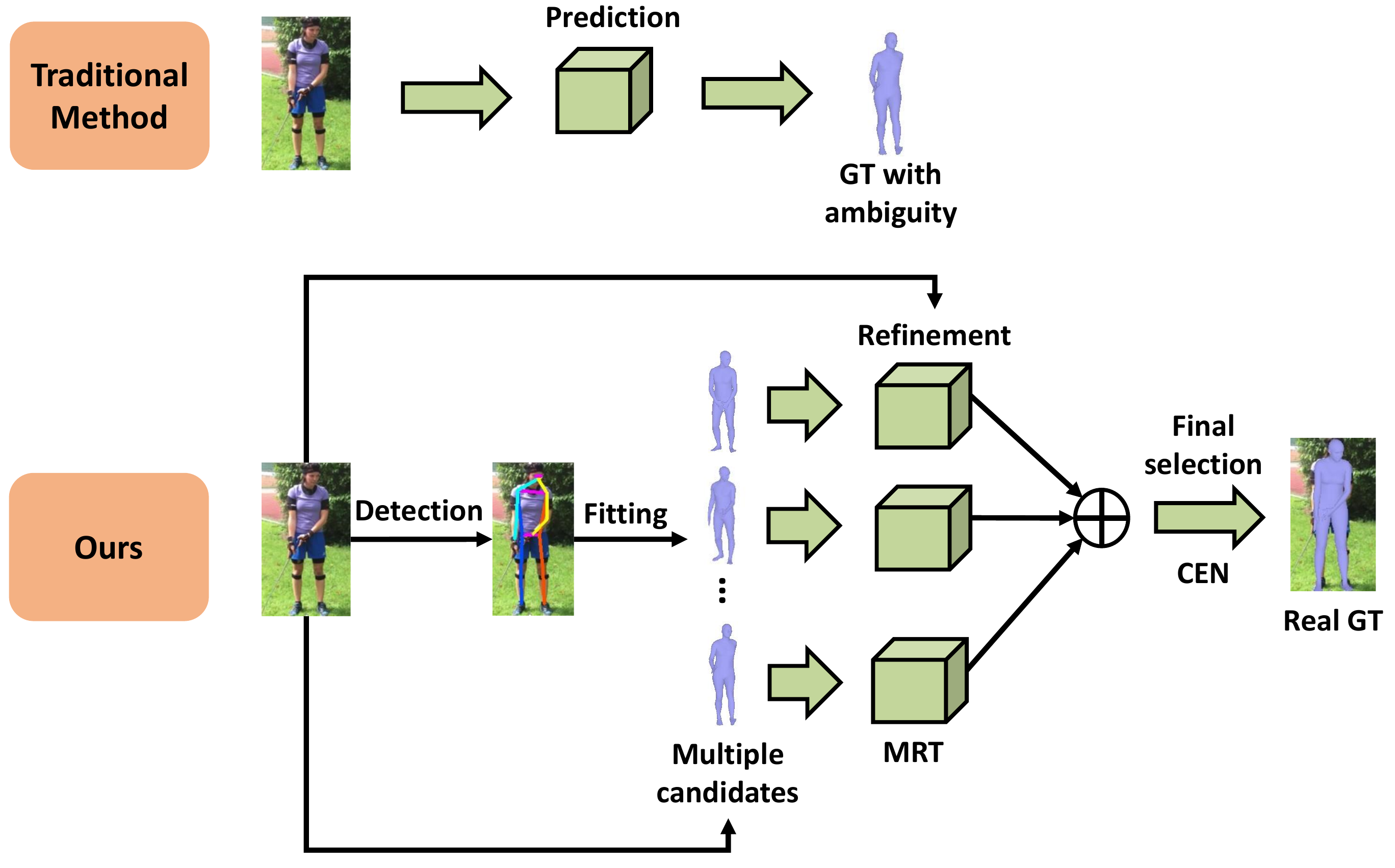}
		\end{minipage}
	}
	\caption{The problem caused by weak supervision label and our solution}
	\label{fig:fig_compare}
\end{figure}

\section{Introduction}
With the help of the recent developed parametric model of the human body, monocular image 3D human pose and shape reconstruction has achieved great advancements in recent years
~\cite{carion2020end, omran2018neural, kolotouros2019learning, kolotouros2019convolutional}. 
Nowadays optimization-based and regression-based approaches are two representative 
reasearch branches of this field. Optimization-based approach optimizes model parameters by 
fitting the human body model to 2D keypoints or other kinds of weak supervision labels, e.g. part segementation~\cite{yang2020renovating}, densepose~\cite{guler2018densepose, yang2019parsing}. However,  
optimizing these weak labels inevitably suffers from 2D-to-3D ambiguity problem. 
For example, 2D observations lack depth information, 
and 3D keypoints lack the information of rotation angle on limb axis. Therefore, optimization-based approaches 
are sensitive to the choice of initialization and easy to converge into a local optimum or unrealistic result. 

Regression-based approaches apply a deep CNN to regress 
the 3D human pose and shape parameters from a RGB input image. Due to the difficulty of in-the-wild 
3D training data acquisition, existing regression based methods also heavily rely on the supervision of weak 
labels, including 2D projection loss~\cite{lin2014microsoft}, 
3D keypoints loss~\cite{ionescu2013human3}, body silhouette loss~\cite{yao2019densebody} and densepose loss
~\cite{rong2019delving}. Since each weak label corresponds with many possible 3D bodies, 
network with a single output cannot 
always select the real 3D reconstruction from different possible results to match with the weak 
supervision label, leading to the unsatisfied training quality.

In general, both two kinds of approaches suffer from the ambugity problem caused by the weak supervision. 
In this paper, we aim to solve this issue and discover the most 
accurate result from multiple possible 3D reconstructions which are matched with the same weak label.   
Instead of using a single network that outputs a single reconstruction result in one propagation, 
we decompose the internal analysis process of network and propose a novel three-stage framework named 
Multi-Initialization Optimization Network (MION) to predict the most appropriate reconstruction 
result through multiple optimizition branches. 

Specially, in the first stage, we design a reconstruction candidate optimization strategy to optimize 
several different coarse reconstruction candidates for each sample. To achieve this, we first generate a 
human mesh candidate pool by clustering a big human motion capture dataset combination~\cite{mahmood2019amass}. 
Then for each candidate in the pool, we optimize its camera parameters by fitting the 3D mesh model to a 
weak observation of the given sample. After this, the optimized candidate with  
a relative small fitting loss can be selected to get into the next inference stage and regarded as an initialization 
to start one optimization branch. Moreover, we accelerate the parameter optimization process by directly computing its 
closed-form solution. Based on the multiple candidates, our method has more possibilities to avoid the local optimal 
and find the real 3D reconstruction. 

In the second stage, we continue to push each optimization branch forward and 
design a mesh refinement transformer (MRT) to refine each coarse reconstruction candidate. 
This transformer framework has two advantages. Firstly, to achieve a refinement process, 
the initial 3D reconstruction information from the last stage can be effectively encoded into 
the transformer by a novel Projected Normalized Coordinate Code (PNCC) positional encoding. 
Secondly, transformer can adaptively learn the non-local relationships between different body 
joints during training stage, which is significant for enhancing the body structure prediction ability.

Finally, the refined 3D reconstruction candidates from different optimizition branches need to be aggregated to 
one result. In the last stage, a Consistency Estimation Network (CEN) is proposed to 
distinguish if the form of each 3D reconstruction matches the visual evidence in the input image 
and select the best 3D reconstruction result. 
In order to get rid of the limitation of label ambiguity, CEN benefits from 
a specific data synthesis strategy which generates the accurate 3D ground-truth for each training sample. 
The overview of our Multi-Initialization Optimization Network (MION) is shown in Fig.~\ref{fig:fig_compare}.

Our contributions can be summarized as follows:
\begin{itemize}
	\item In order to deal with the issue that weak supervision labels have ambiguities. 
	We propose a novel three-stage framework named Multi-Initialization Optimization Network (MION) to 
	predict appropriate human pose and shape reconstruction through multiple optimizition branches.
	\item In our MION, instead of predicting a single result, for each sample, 
	we calculate mutiple coarse reconstruction candidates as the initalizations to start different optimization branches. Compared with a single prediction, multiple candidates give more possibilities to avoid the local optimal and find the real 3D reconstruction.
	
	\item Given the different initial reconstruction candidates, we design a mesh refinement 
	transformer (MRT) with a novel Projected Normalized Coordinate Code (PNCC) positional encoding 
	to further refine each coarse reconstruction via a self-attention mechanism. 
	Then a Consistency Estimation Network (CEN) is proposed to select the best 3D reconstruction result 
	from all the optimization branches.
	
	\item Both qualitative and quantitative experiments show that our MION significantly improves 
	the performance of monocular image 3D human reconstruction and achieves the state-of-the-art result among 
	other methods.  
	
\end{itemize}

\section{Related works}
\textbf{Optimization based methods. }
With the developement  of parametric 3D human body model, such as SCAPE~\cite{anguelov2005scape}, SMPL~\cite{loper2015smpl}, optimization based 3D 
human reconstruction method becomes an important branch in the research community. This kind of method infers 3D reconstruction by 
fitting a parametric model to match the given 2D observation. Early optimization methods~\cite{agarwal2005recovering, sigal2007combined} 
fit the human body model by the manually generated keypoints and silhouettes labels. These methods rely on manual 
intervention and generalize badly to images in the wild. Federica \textit{et al.}~\cite{bogo2016keep} propose 
first automatic 3d human model 
reconstruction method SMPLify which fits SMPL model to the 2D keypoints predicted by CNN detector~\cite{pishchulin2016deepcut}. 
Meanwhile, objection function contains different regularization terms to ensure the optimization can produce a plausible result. 
In order to further improve the performance, more supervision information are incorporated into the optimization target, such as silhouette~\cite{lassner2017unite}, scene constrains~\cite{zanfir2018monocular} and multi-view~\cite{huang2017towards}. 

Generally, these optimization based methods are sensitive to the choice of initialization and tends to have a slow optimization speed. 
Meanwhile, the optimization process only converges to one local optimal result by the 2D observation input without appearence information. 
Thus it suffers from the problem of label ambiguities.

\textbf{Learning based methods. }
Another representative method is learning base reconstruction method, which learns a human model parameter regressor by a data-driven way. 
Because of the lack of in-the-wild 3D reconstruction paired training data. Existing methods focus on the weak supervision way to solve this 
issue. HMR~\cite{kanazawa2018end} directly regresses SMPL parameters from images by a CNN and 
adds iterative regression to further improve the accuracy. 
It also proposes an adversarial prior in case the reconstruction is not realistic. SPIN~\cite{kolotouros2019learning} incorporates a 
optimization into the network learning process, where the predicted parameter is refined by a optimization process to further 
supervise the network. These two methods both apply 2D and 3D keypoints as weak supervision signal.  
Inspired by the dense correspondence representation used in DensePose~\cite{guler2018densepose}, various 
learning based methods~\cite{guler2019holopose, rong2019delving} regard IUV map as intermediate 
representation or weak supervision label for improving the regression CNN. 
On the other hand, in order to remove the limitation of SMPL parameter space, many learning based methods do not rely on the parametric 
model and directly regress the coordinates of each vertices on the mesh.  BodyNet~\cite{varol2018bodynet} regresses a 
volumetric representation of 3D human by a Voxel-CNN. Densebody~\cite{yao2019densebody} and DaNet~\cite{zhang2019danet} use a UV position 
map to represent 3D human body. 

All the aforementioned learning based methods rely on weak supervision label during tranining stage. It is flawed in that the label 
with ambiguites cannot always lead the network to predict the real reconstruction. In this work, with the same weak supervision labels,  
we propose a multi-path optimization based reconstruction framework to reduce the possibility of getting into local optimal. 
In order to deal with the occlusion cases with ambiguites, Biggs \textit{et al.}~\cite{biggs20203d} also proposes to predict a candidate set which contains different possible reconstruction results. However, their network is only trained on the single Human3.6m dataset~\cite{ionescu2013human3} 
where each sample already has 3D reconstruction ground-truth. Thus it still cannot solve the problem raised by the weak label. 

\textbf{Synthetic data. }
Since it is diffcult to collect in-the-wild samples with 3D ground-truth, synthetic data plays an important role in 3D human pose and shape 
estimation task. Pavlakos \textit{et al.}~\cite{pavlakos2018learning} adopt joint heatmap and silhouette as intermediate representation and design 
two decoders to respectively predict the pose and shape parameter from the above two representations. Without introducing appearance 
information, the decoders can be trained by synthetic data. Xu \textit{et al.}~\cite{xu2019denserac} propose a network to decode human body from 
a synthetic IUV map. Although the above methods benefit from the abundant synthetic training data, the inference process from 
weak label to 3D ground-truth is an ambiguity task which cannot be solved by even human. 

Existing synthetic data based methods usually ignore the appearance information. Different from them, we utilize synthetic data to help 
the network discriminate if the visual evidence in RGB image matches a given 3D reconstruction, which can be used to select the best 
result from multiple candidates. 

\textbf{Human structure dependence. }
Human body structure in natural world has a strong prior, which builds the dependencies between different parts of the whole human body. 
Some existing methods explict model such dependencies in their learning framework. CMR~\cite{kolotouros2019convolutional} apply a 
Graph-CNN~\cite{kipf2016semi} to model the interactions between different vertices in the inference framework. 
METRO~\cite{lin2020end} replaces Graph-CNN with a Transformer encoder~\cite{vaswani2017attention} to model the interactions. 
However the query in its Transformer only contains the global feature of the input image and abandons detailed local information. 
In this work, we apply a full encoder-decoder Transformer structure which maintains the local image appearance information to refine the body 
structure.

\section{Method}
As above mentioned, the whole framework of our Multi-Initialization Optimization Network (MION) has 
three stages. In this section, we explain the details of each stage in sequence.  

\subsection{reconstruction candidate selection}

For common traditional CNN regression based 3D human reconstruction frameworks, training with 
weak supervision labels might make the network fall into local optimum. 
As long as the predicted 
3D reconstruction matches with the provided weak supervision labels, for e.g., the 2D landmarks or masks. 
the network training will stop no matter if the current prediction is correct. 
In this work, we aim at designing a new three-stage framework begins with multiple initializations.

At our first stage, considering that nowadays 2D body keypoints telenology already achieves a high performence, 
For each training sample, we first utilize HRNet~\cite{sun2019deep} to detect its 2D body keypoints which 
represents the weak supervision label. Then we expect to coarsely locate all the representative 
possible solutions according to this 2D keypoints 
in the whole solution space. To this end, a candidate selection strategy is proposed to calculate 
multiple possible 3D reconstruction candidates for each sample. Each candidate can be regard as an inital 
optimization point.  

\begin{figure}[t]
	\begin{center}
		\includegraphics[width=\linewidth]{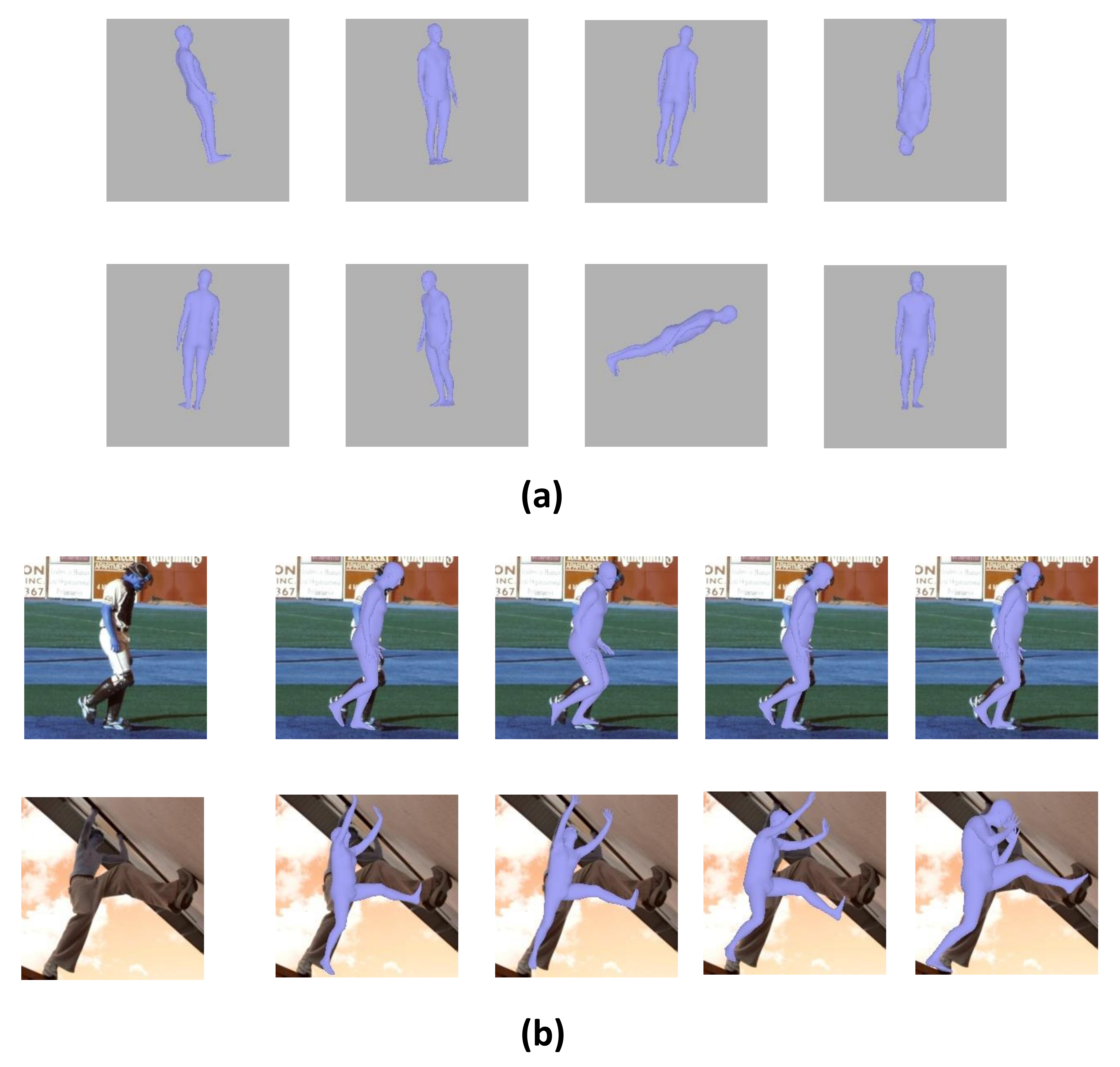}
	\end{center}
	\caption{(a) Visualization of camera orientation cluster result. (b) Visualization of the selected 
		candidates}
	\label{fig:cluster_vis}
\end{figure}

Specifically, we regard a huge human motion capture dataset named AMASS~\cite{mahmood2019amass} as the source 
of our candidate and assume it contains all the possible human poses shown in natural images. Each 3D 
human mesh sample in AMASS is expressed by a set of SMPL~\cite{loper2015smpl} parameters. SMPL is a 
parametric model for human body mesh representation, which maps the shape 
parameters $\beta \in R^{10}$ and the pose parameters $\theta \in R^{72}$ to the human body mesh 
$ M \in R^{V \times 3}$ by a linear model. Based on this, 
to generate the reconstruction candidates of a 
specific sample, we first optimize a perspective projection matrix for each candidate 
in AMASS by fitting the 3D keypoints from candidate meshes to the 2D keypoints from the given sample. 
Then the candidate with a low fitting loss value can be 
selected as an possible solution to get into the next inference stage.

However, AMASS is a huge candidate dataset. It is unrealistic to optimize all the perspective projection 
matrix parameters for each candidate because of the high computational cost. 
In order to reduce the cost, we calculate the representative candidate bodies from the whole AMASS 
dataset by k-means clustering. Then the approximate 
optimal solution can be directly selected from the cluster centroids. In this work, we set the number of 
pose parameter clusters to 10000 and the number of camera orientation parameter clusters to 30, making the 
whole candidate pool contain 300000 members. Fig~\ref{fig:cluster_vis}(a) visualizes different 
cluster centroids of camera orientation parameter. 
The optimization target function can be defined as:

\begin{equation}
\max_{T}\mathcal{L} = \|\Pi_{K}(J_{cand}^{3d}, T) - J_{gt}^{2d}\|^2_2\\
\end{equation}
\begin{equation}
\Pi_{T}(J_{cand}^{3d}, T) =
\begin{bmatrix}
f&0&c_1&\\
0&f&c_2&\\
\end{bmatrix}
\begin{pmatrix}[1.8]
\frac{J_x + T_x}{J_z + T_z}&\\
\frac{J_y + T_y}{J_z + T_z}&\\
1 &\\
\end{pmatrix}
\end{equation}
where $J_{cand}^{3d} \in R^{N \times 3}$ denotes the cordinates of a set of 3D keypoints regressed from a candidate mesh, 
and $J_{gt}^{2d} \in R^{N \times 2}$ is the corresponding 2D keypoint. $\Pi_{K}$ is the 
perspective projection function from 3D to 2D. $T$ is camera translation parameters for optimizing. 
$f$ is the pre-defined focal length of camera, $c_1$ and $c_2$ are the pre-defined camera center 
parameters. 
Since the camera orientation parameters and pose parameters are already known, only the three camera translation parameters are required to be optimized. 
To further accelerate the computing speed, we directly compute the closed-form solution of the ternary homogeneous linear equations. This 
implementation omits the time cost on Gaussian elimination compared with the traditional least square method. Based on this, our candidate 
selection process only takes 18 ms on GPU. 

After fitting all the 3D candidates to the given 2D keypoints, 
we select a set of candidates with low fitting loss and a large pose variance, although the 
selected candidates match with the same 2D keypoints, they still might distribute dispersedly 
in the pose solution space to give more initializations for avoiding the local optimal. The selected 
reconstruction candidates of two samples is shown in Fig.~\ref{fig:cluster_vis}(b). 

\begin{figure}[t]
	\begin{center}
		\includegraphics[width=\linewidth]{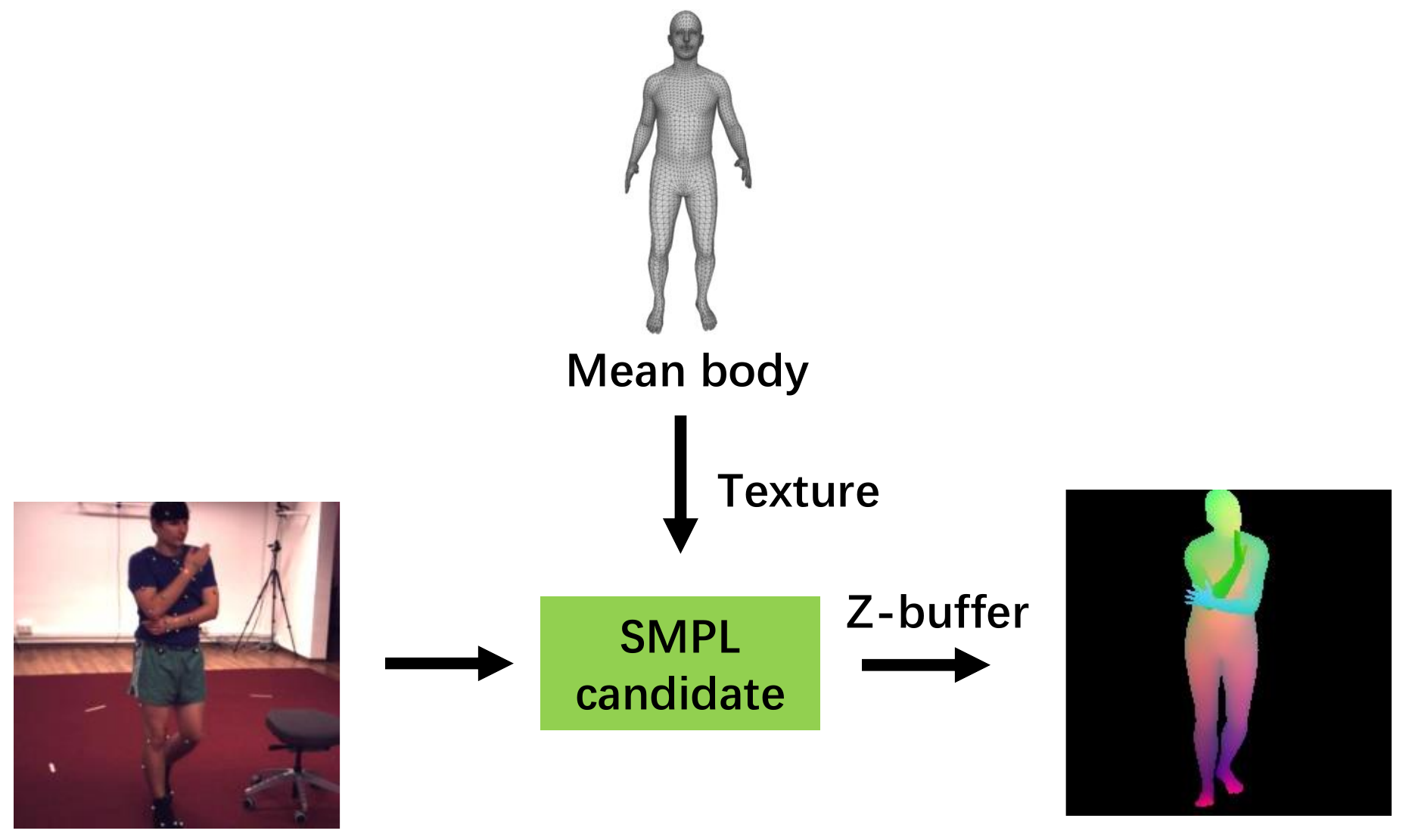}
	\end{center}
	\caption{The process of PNCC generating}
	\label{fig:pncc}
\end{figure}

\subsection{Mesh Refinement Transformer}
Most selected candidates from the first stage are corasely optimized and need further refinement. 
In order to alleviate local optimal solutions and increase the possibility of finding the real 
3D reconstruction, in this stage, we provide an individual optimization branch to refine each coarse candidate. 
Each branch only focuses on refining the details to 
make the initialization more compatible with the weak supervision label.

\begin{figure*}
	\begin{center}
		\includegraphics[width=\textwidth]{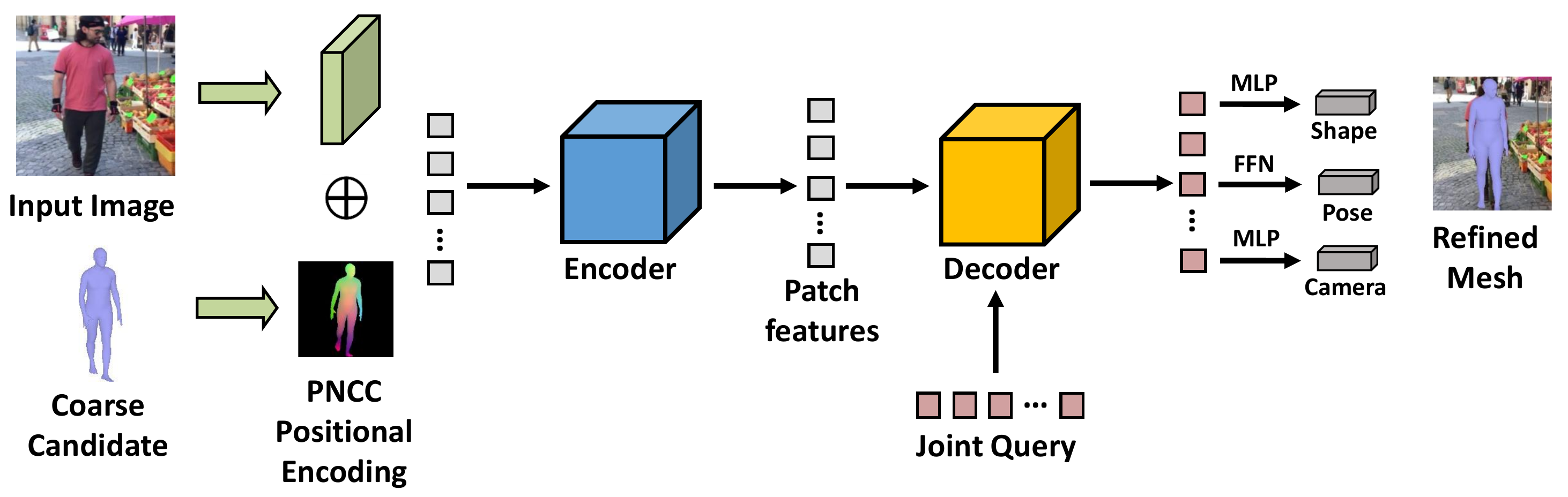}
	\end{center}
	\caption{The framework of Mesh Refinement Transformer}
	\label{fig:crt}
\end{figure*}

Inspired by a series of recent vision transformer (ViT) works~\cite{carion2020end, dosovitskiy2020image, chen2020pre, zheng2020rethinking}, 
we determine to use the ViT architecture which has two advantages for 
the body reconstruction refinement task. Firstly, ViT follows a sequence prediction format by regarding the input image as a sequence of different local patches. This manner allows the whole inference framework pay attention on the details of each local patch, which is suitable for the refinement task. Secondly, the self-attention mechanism of transformers, which explicitly models all pairwise interactions between elements in a sequence makes our architectures particularly suitable for learning the relationship of different body parts.

To construct our transformer network, we first use a backbone network (e.g. resnet) to extract 
image feature~\cite{chen2020pre}. Then three deconvolution layers are added to the top layer of backbone 
to upsample the feature map and recover more spatial information. 
Finally we flatten feature map to make a feature vector sequence and input it into a transformer encoder, as 
shown in Fig.~\ref{fig:crt}. Thus each element in the sequence represents a local patch 
of input image.  

In order to refine the SMPL parameter of a given candidate. we need to encode the candidate 
as a 2D map for CNN, which corresponds to the positional encoding in the common transformer. To this end, inspired by 
Projected Normalized Coordinate Code (PNCC) used in 3D face alignment work 3DDFA~\cite{zhu2016face}, we propose a Projected 
Normalized Coordinate Code based positional encoding (PNCC-PE) which 
encodes the initial SMPL parameter information into the transformer input. 
Specifically, we first normalize the mean body point cloud to 0-1 in $x$, $y$, $z$ axis as Eq. (\ref{Eq:NCC})
\begin{equation}\label{Eq:NCC}
\text{NCC}_{d}=\frac{\mathbf{\overline{S}}_{d} - \min(\overline{\mathbf{S}}_{d})}{\max(\overline{\mathbf{S}}_{d}) - \min(\overline{\mathbf{S}}_{d})}~~~ (d = x,y,z),
\end{equation}
where $\mathbf{\overline{S}} \in R^{V \times 3}$ is the point cloud tensor of mean SMPL model. We regard the 
three channel normalized coordinate code (NCC) of each vertice as its texture. Then 
PNCC is generated by adopting Z-Buffer to render the body mesh with inital SMPL parameter on a zero 
value background, as shown in  Eq. (\ref{Eq:PNCC})
\begin{equation}\label{Eq:PNCC}
\text{PNCC}= \emph{Z-Buffer}(\mathcal{M}(\theta_{cand}, \beta), \gamma_{cand}, \text{NCC})
\end{equation}
where $\theta_{cand}$ and $\gamma_{cand}$ are the pose and camera parameters of a candidate predicted by RCG. 
All the candidates apply a mean SMPL shape parameter $\beta$. 
The PNCC calculation process is shown in Fig.~\ref{fig:pncc} 
After we get PNCC map, the sine and cosine functions used in original 
transformer positional encoding~\cite{vaswani2017attention} are applied to transfer the 3-channel PNCC 
into the final positional encoding feature (PNCC-PE). Each value on the PNCC-PE can be represented by: 
\begin{equation}\label{Eq:postional}
\begin{split}
PNCC-PE_{(pos, 2i)} = sin(PNCC_{(pos)} / 10000^{2i/d_{model}})\\
PNCC-PE_{(pos, 2i + 1)} = cos(PNCC_{(pos)} / 10000^{2i/d_{model}})\\
\end{split}
\end{equation}
where $pos$ is denotes the spatial position index on PNCC map, $i$ is the channel index of PNCC-PE. PNCC-PE and 
backbone feature have the same channel dimension number. Thus they can be added and input to the transformer encoder. 
Based on this, PNCC-PE builds the relationship between each local image patch and 
its corresponding body part by the provided prior information of candidate. 
This design helps the tranformer pay more attention to the useful local patch details when refining each body parts. 

The encoder of Mesh Refinement Transformer 
(MRT) follows the standard multi-head attention and feed-forward networks architecture. 
Since the pose parameter of a SMPL model is composed of a set of joint rotation vectors, we decompose the pose 
representation and make each input query of our decoder represent the embedding of one joint rotation vector. 
Then the output sequence of decoder is the refined pose parameter. 
Meanwhile, we add two MLP networks which respectively predict the camera parameter and shape parameter from the 
top sequence feature. 
Therefore, all the weak supervision labels (e.g. 2D keypoints) 
can be involved into the training by a perspective projection. The overview of our MRT is shown in Fig.~\ref{fig:crt}

The total loss of our MRT is as follows:
\begin{equation}\label{Eq:loss_mrt}
\mathbf{L} = w_1 * L_{smpl} + w_2 * (L_{J}^{3d} + L_{J}^{2d})
\end{equation}
\begin{equation}\label{Eq:loss_smpl}
L_{smpl} = \|[\theta_{reg}, \beta_{reg}] - [\theta_{gt}, \beta_{gt}]\|_{2}
\end{equation}
\begin{equation}\label{Eq:loss_3d}
L_{J}^{3d} = \|J_{gt}^{3d} - R(V_{reg})\|_{2}
\end{equation}
\begin{equation}\label{Eq:loss_2d}
L_{J}^{2d} = \|J_{gt}^{2d} - \Pi_{K}(R(V_{reg}))\|_{2}
\end{equation}
where $L_{v}$, $L_{J}^{3d}$ and $L_{J}^{2d}$ are the vertice loss, 3D joint loss and 2D joint projection loss. 
$w_1$ and $w_2$ are the weights of different loss functions. 

\subsection{Consistency Estimation Network}
Applying multiple initalizations usually leads to different optimized results. Some results might fall into 
local optimum and some results might be close to the real 3D ground-truth. Therefore, it is necessary to find the best reconstruction 
candidate among all the refined candidates. To this end, in the last stage, we consider the optimum path selection as a 
scoring problem and propose a Consistency Estimation Network (CEN) to solve this problem by utilizing the synthesis 
training data SURREAL~\cite{varol2017learning}.

The target of the network is to identify 
if one 3D reconstruction matches the visual evidence of input human image. For this purpose, 
we need a dataset with ground-truth 3D body mesh, which can be achieved by data synthesis.
Specifically, when generating a synthesis sample, we select one training sample and randomly pick two candidates from 
its candidate collection generated in the first stage. 
Then we use the SMPL parameter of one candidate to render a RGB synthesis image. Following the synthesis 
method of SURREAL~\cite{varol2017learning} dataset, the body texture are selected from 
its own texture set and the background image is selected from a subset of LSUN dataset~\cite{yu2015lsun}.

\begin{figure}[t]
	\begin{center}
		\includegraphics[width=\linewidth]{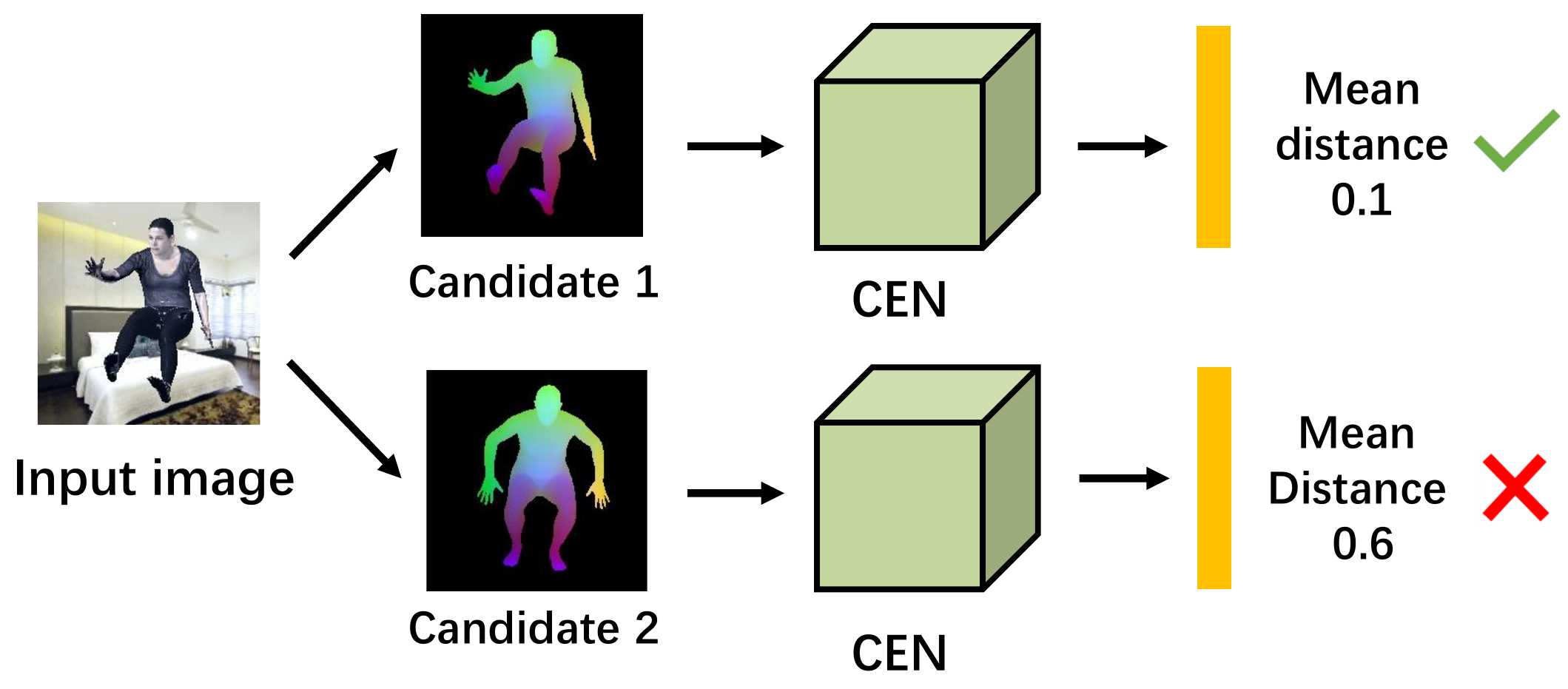}
	\end{center}
	\caption{The inference process of Consistency Estimation Network.}
	\label{fig:arn}
\end{figure}

The other picked candidate is used to simulate a predicted result from the last stage. Following the same way as MRT, we use its SMPL parameters to render a PNCC map to represent the reconstruction information. Finally, as shown in Fig.~\ref{fig:arn}, PNCC is concatenateed 
with the input image and sent to a CNN which regresses the vertice wise distance of the two picked candidates. 
The loss function of our Consistency Estimation Network (CEN) is as follows:
\begin{equation}\label{Eq:loss_arn}
L_{i} = \|P_{reg}^{i} - \|V_{gt}^{i} - V_{cand}^{i}\|_{2}\|_{2}
\end{equation}
where $i$ is the index of one vertice on the body point cloud, $P_{reg}^{i}$ is the $i$th output of our CEN, 
meaning the regressed score of $i$th vertice. 
$\|V_{gt}^{i} - V_{cand}^{i}\|_{2}$ denotes the distance between the PNCC encoding body and RGB image encoding body 
in a normalized point cloud space. During the inference stage, we 
compute the average distance of all the output vertice distance and choose the reconstruction result in the optimization branch with the 
lowest distance as our final result.

\subsection{Implementation details}
For our candidate selection strategy in the first stage, after we fit all the candidates to the 2D keypoints 
of current sample, all the candidates with a fitting loss lower than 2000 are choosen to be an avaliable candidate. 
Then we iteratively select the candidate which has the largest pose parameter distance with the selected candidates 
and put it into the final candidate collection. 

For the Mesh Refinement Transformer (MRT), the backbone network adopts the architecture 
of ResNet-50~\cite{he2016deep}. Note that we remove the last fully connection layers in original ResNet-50 and add three deconvolution layers to make a fully convolution network (FCN) as our backbone. The FCN receives the 224×224
input image and produces 56 × 56 feature maps with 384 channels. In order to match with image feature, the rendered 
PNCC has the same resolution of 56 × 56 and each channel of PNCC is transfered into a position encoding map with 128 channel. During MRT training stage, the loss weight of SMPL parameter regression is set to 1 and the loss weight 
of 2D keypoints and 3D keypoints regression is set to 5. The data augmentation techniques includes 
rotation $[-60^{\circ} , 60^{\circ}]$ color jittering $[0.6 , 1.4]$ and flipping, are
applied randomly to input images. We adopt the AdamW~\cite{loshchilov2017decoupled} optimizer with 
an initial learning rate of $5 \times 10^{-5}$ to train the MRT model, and reduce the learning
rate to $5 \times 10^{-6}$ after 20 epochs. The training process stops after 60 epochs. MRT is trained on 4 
Titan X GPUs with a batch size of 64. 
During training and inference stage, all the 2D keypoints used in the first stage are predicted by 
the HRNet-W48 network in MMPose toolbox~\cite{mmpose2020}.

For training the Consistency Estimation Network (CEN), we adopts a ResNet-34 network 
as backbone. The inital learning rate is set to 0.01. We train our model for 100 epochs and 
lower the learning rate by a factor of 10 after 50 epochs.

\section{Experiments}
This section focuses on the empirical evaluation of the
proposed method. First, we present the datasets and evaluation metrics that
we employed for quantitative and qualitative evaluation.
Then, we conduct extensive ablation experiments and comparsion experiments to verify the 
effectiveness of our method.

\subsection{Datasets and Evaluation Metrics}

The networks mentioned in this work are trained on various datasets. Specifically, the training sets of our 
Mesh Refinement Transformer (MRT) is dominated by weak supervision datasets, including Human3.6M~\cite{ionescu2013human3}, 
LSP~\cite{johnson2010clustered}, MPII~\cite{andriluka20142d}, COCO~\cite{lin2014microsoft}, 
LSP-Extended~\cite{johnson2011learning}, MPI-INF-3DHP~\cite{mehta2017monocular}. Our 
Consistency Estimation Network (CEN) is 
trained on UP-3D~\cite{lassner2017unite} and SURREAL~\cite{varol2017learning} which have 3D grouth-truth. We conduct the evaluations 
on the test set of Human3.6M and 3DPW~\cite{von2018recovering}. To get a pair comparison with earlier state-of-the-art method, we use the 
same evalution metric with SPIN~\cite{kolotouros2019learning}.

~\textbf{Human3.6M}: It is an indoor benchmark for 3D human pose estimation. It includes multiple subjects performing actions like Eating, Sitting and Walking. Following typical protocols, e.g., ~\cite{kolotouros2019learning}, we use subjects S1, S5, S6, S7, S8 for training and we evaluate on subjects S9 and S11. 

~\textbf{LSP}: LSP with its extension is a standard 2D human pose estimation dataset which 
is collected by the images from sports activities. This dataset has large variance in terms of appearance 
and especially articulations. Each person in this dataset is labeled with total 14 joints which are used for weak 
supervison by the perspective projection. 

~\textbf{MPII}: MPII is a 2D human pose estimation dataset which covers a wide range of human activities
with 25k images containing over 40k people. We use it for weak supervision during training stage. 

~\textbf{MPI-INF-3DHP}: It is a dataset captured with a multi-view setup mostly in indoor environments. No markers are used
for the capture, so 3D pose data tend to be less accurate compared to other datasets. We use the provided training
set (subjects S1 to S8) for training. 

~\textbf{UP-3D}: It is a recent dataset that collects color images from
2D human pose benchmarks. SMPLify~\cite{bogo2016keep} is utilized to generate 3D human shape candidates for each sample. The candidates were 
evaluated by human annotators to select only the images with good 3D shape fits. It comprises 8515 images, where 7818 are used for training. 

~\textbf{SURREAL}: It provide a tool to generate synthetic image examples with 3D shape ground truth. In this work, we select the 
SMPL pose parameters from AMASS and select the background images from LSUN dataset~\cite{yu2015lsun}. Each training sample is generated by a 
common rendering pipeline and has an accurate 3D ground-truth. 

~\textbf{Evaluation Metrics}: 
Follwing previous method~\cite{zhang2019danet, guler2019holopose, pavlakos2019texturepose}, the evalution is conducted by two popular protocols: Mean Per Joint Position Error (MPJPE) and the MPJPE after rigid alignment 
of the prediction with ground truth using Procrustes Analysis (MPJPE-PA). Both two metics measures the 
Euclidean distances between the ground truth joints and the predicted joints.

\subsection{Ablation experiment}

To evaluate the effectiveness of each component proposed in our
method, we conduct ablation experiments on Human3.6M under
various settings.

~\textbf{Effect of different number of initalizations} In order to study the effect of using 
different number of initalizations in our 3D human reconstruction framework, 
we respectively test our Multi-Initialization Optimization Network (MION) under 
different numbers of optimization initalizations. 
As shown in Tab.~\ref{table_path_number}, when only using one initialization to inference the reconstruction, 
our MPTN degrades to a single Mesh Refinement Transformer (MRT). Not surprisingly, one initialization method 
leads to the worst result. When applying more candidates into the whole framework, the performance is 
continuously improved until the number of candidates reaches 5.  This phenomenon indicates that 
applying multiple different optimization initalizations is an effective way to alleviate 
falling into local optimum and improve the overall performence in 3D human reconstruction task

\begin{table}[t]
	\centering
	
	\begin{tabular}{lcc}
		\toprule
		Method   & MPJPE & PA-MPJPE \\
		\midrule
		1-path (MRT) & 62.31 & 45.48  \\
		2-path (MION)& 61.59 & 44.75 \\
		3-path (MION)& 59.98 & 42.81 \\
		4-path (MION)& 58.78 & 41.86 \\
		5-path (MION)& \textbf{56.88} & \textbf{41.59} \\
		6-path (MION)& 58.71 & 42.02 \\
		\bottomrule
	\end{tabular}
	\vspace{1em}
	\caption{Analysis the difference of using different number of optimization paths. 
		MPJPE and PA-MPJPE are used as evaluation metric.}
	\label{table_path_number}
\end{table}

~\textbf{Effectiveness of PNCC position encoding} 
To demonstrate the superiority of our Projected Normalized Coordinate 
Code based positional encoding (PNCC-PE) over traditional position encoding method, 
we further conduct experiment to investigate the impact of PNCC-PE. 
Tab.~\ref{table_pe} shows the ablation study on Human3.6M. 
MION (w/o PNCC-PE) replaces the PNCC-PE by the traditional sinusoidal position encoding~\cite{vaswani2017attention}. 
As we can see, compared with traditional position encoding, the proposed PNCC-PE significantly improve the 
performence on 3D human reconstruction task: 56.88\% vs 61.14\%. We attribute this phenomenon to the 
initalization information brought by the PNCC-PE makes the joint queries of decoder easily focus on the 
corresponding local patch feature. 

\begin{table}[t]
	\centering
	
	\begin{tabular}{lcc}
		\toprule
		Method   & MPJPE & PA-MPJPE \\
		\midrule
		MION w/o PNCC-PE & 61.14 & 43.01  \\
		MION  & \textbf{56.88} & \textbf{41.59} \\
		\bottomrule
	\end{tabular}
	\vspace{1em}
	\caption{Analysis of the effective of using the PNCC position encoding 
		in a transformer framework. MPJPE and PA-MPJPE are used as evaluation metric.}
	\label{table_pe}
\end{table}

~\textbf{Effectiveness of Consistency Estimation Network} 
Consistency Estimation Network (CEN) is one of the key steps in our algorithm. 
To demonstrate the effectiveness of our CEN on selecting the best optimized result, 
we make a performence comparison between different 3D reconstruction selection methods in the third stage. 
First, we apply the random selection strategy instead of CEN. As indicated in 
Tab.~\ref{table_arn}, compared with random selection strategy MION w/o CEN, our CEN reduces the MPJPE error from 
62.47\% to 56.88\%,  which proves the effectiveness of CEN on predicting the consistency between 
RGB image and given human body parameter. We also evaluate the upper bound of CEN by always selecting the 
result with lowest error. As we can see, the performence gets further improvement: from 56.88\% to 52.17\%, 
meaning that the multiple initialization method has the potential to achieve a better result. 

\begin{table}[t]
	\centering
	
	\begin{tabular}{lcc}
		\toprule
		Method   & MPJPE & PA-MPJPE \\
		\midrule
		MION w/o CEN & 62.47 & 44.71  \\
		MION with CEN   & \textbf{56.88} & \textbf{41.59} \\
		MION best path  & 52.17 & 39.82 \\
		\bottomrule
	\end{tabular}
	\vspace{1em}
	\caption{Analysis of the effective of using Consistency Estimation Network (CEN) 
		in a transformer framework. MPJPE and PA-MPJPE are used as evaluation metric.}
	\label{table_arn}
\end{table}

\subsection{Comparison experiment}

\textbf{Comparison on the In-door Dataset.}

\begin{table}[t]
	\centering
	
	\begin{tabular}{lcc}
		\toprule
		Method   & MPJPE & PA-MPJPE \\
		\midrule
		SMPLify~\cite{bogo2016keep}   & - &  82.3  \\
		NBF~\cite{omran2018neural}   & - & 59.9 \\
		HMR~\cite{kanazawa2018end}   & 88.0 & 56.8  \\
		GraphCMR~\cite{kolotouros2019convolutional}   & - & 50.1  \\
		HoloPose~\cite{guler2019holopose}    & 64.3 & 50.6  \\
		TexturePose~\cite{pavlakos2019texturepose}   & - & 49.7  \\
		DaNet~\cite{zhang2019danet}          & 61.5 & 48.6  \\
		DenseRaC~\cite{xu2019denserac}          & 76.8 & 48.0  \\
		Pose2Mesh~\cite{choi2020pose2mesh}   & 64.9 & 47.0  \\
		SPIN~\cite{kolotouros2019learning}   & 62.3 & \textbf{41.1}  \\
		\midrule
		MION & \textbf{56.88} & \textbf{41.59} \\
		\bottomrule
	\end{tabular}
	\vspace{1em}
	\caption{Comparison with state of the art on Human3.6M dataset. MPJPE and PA-MPJPE are used as evaluation metric.}
	\label{table_compare-h36m}
\end{table}

We evaluate the performance of our methods on the in-door Dataset Human3.6M in terms of 3D pose estimation accuracy. 
We train our model following the setting of SPIN~\cite{kolotouros2019learning} and utilize 
Human3.6M, LSP, MPII, COCO and MPI-INF-3DHP as the training set. 
Quantitative results are reported in Tab.~\ref{table_compare-h36m}. It shows the results of our approach 
against other the state of the art methods which output a full mesh of the human body (SMPL, in particular). 
As we can see, training with the same amount of weak supervision label, our method significantly 
outperforms other methods on MPJPE metric and achieves a competitive performence on PA-MPJPE metric.

\begin{table}[t]
	\centering
	
	\begin{tabular}{lcc}
		\toprule
		Method   & MPJPE & PA-MPJPE \\
		\midrule
		HMR~\cite{kanazawa2018end}   & - & 81.3  \\
		GraphCMR~\cite{kolotouros2019convolutional}   & - & 70.2 \\
		STRAPS~\cite{sengupta2020synthetic}   & - & 66.8 \\
		SPIN~\cite{kolotouros2019learning}   & - & 59.2  \\
		Pose2Mesh~\cite{choi2020pose2mesh}   & 89.2 & 58.9  \\
		I2LMeshNet~\cite{moon2020i2l} & 93.2 & 57.7  \\
		Song\textit{et al.}~\cite{song2020human} & - & 55.9  \\
		\midrule
		MION & \textbf{81.98} & \textbf{52.34} \\
		\bottomrule
	\end{tabular}
	\vspace{1em}
	\caption{Comparison with state of the art on 3DPW dataset. MPJPE and PA-MPJPE are used as evaluation metric.}
	\label{table_compare-3dpw}
\end{table}

\textbf{Comparison on In-the-wild Dataset.}
Since the lack of in-the-wild 3D supervision labels, 
reconstructing 3D human model on in-the-wild outdoor images is much more challenging due
to factors such as extreme poses, appearance variations and heavy occlusions. We conduct evaluation experiments 
on 3DPW datasets to compare our MION with previous 3D human pose and shape estimation methods. As indicated in 
Tab.~\ref{table_compare-3dpw}, 
we can see our method outperforms all the other methods under the 
challenging scenarios, which proves the robustness and generalization of our framework.

\subsection{Analysis experiment}

\textbf{Running speed analysis. }  We evaluate the inference speed of our method and the state of the art 
method SPIN~\cite{kolotouros2019learning} on the same hardware platform (one Titan X Pascal GPU). Both two methods apply the same ResNet50 as the backbone. The whole running time of our MION is 128 ms. Note that our candidate selection process from 300k candidate pool only takes 
18 ms, which is much less than the following CNN inference stages spend (63 ms for the second stage, 47 ms for the third stage). 
In general, since the running time of SPIN is 59ms and we perform much better than SPIN on 3DPW dataset, 52.34 vs 59.2 in 
terms of PA-MPJPE, we believe it is worth taking more time for this better solution. 

\textbf{The performance on shape recovery. } In order to verify the effectiveness of our method on shape recovery, 
we evaluate the shape accuracy of three different methods on 3DPW dataset, including baseline (1-path), MION (5-path) and the 
state of art method SPIN. We apply two evaluation metrics in the experiment, the first one is mean vertex L2 error of two body clouds, 
the second one is the L2 distance between two SMPL shape parameter vectors. The results are shown in Tab.~\ref{table_compare_shape}

\begin{table}[t]
	\centering
	
	\begin{tabular}{lcc}
		\toprule
		Method   & Vertex L2 error & Shape Param L2 error \\
		\midrule
		Baseline (1-path)   & 0.142 &  5.383  \\
		SPIN   & 0.116 & 4.398 \\
		MION (5-path)   & \textbf{0.094} & \textbf{3.227} \\

		\bottomrule
	\end{tabular}
	\vspace{1em}
	\caption{The comparison of different methods in terms of shape recovery accuracy. Mean vertex L2 error and Shape Parameter L2 error 
		are used as evaluation metric.}
	\label{table_compare_shape}
\end{table}

\section{Conclusion} \label{conclusion}

This work aims to solve the 2D-to-3D ambiguity problem of training with weak supervision labels. 
Instead of training a regression network with one output, we propose to apply mutiple initializations and different 
optimization branches to avoid the network easily get stuck in local optimum. Specifically, 
we propose a three-stage framework named Multi-Initialization Optimization Network (MION). 
In the first stage, we strategically select different coarse 3D reconstruction 
candidates which are compatible with the 2D keypoints of input sample.  Regarding each candidate as 
an initialization, in the second stage, we design a mesh refinement transformer (MRT) to respectively
refine each coarse reconstruction result via a self-attention mechanism. Finally, a Consistency Estimation 
Network (CEN) is proposed to find the best result from mutiple candidates by evaluating 
if the visual evidence in RGB image matches a given 3D reconstruction. Experiments demonstrate that our framework 
outperforms existing 3D mesh based methods on multiple public benchmarks.  Future work can focus on improving the efficiency 
of this framework.

\begin{acks}
Acknowledgement: This work was supported by the
Research and Development Projects in the Key Areas of Guangdong Province (No.2019B010153001) and
National Natural Science Foundation of China under Grants No.61772527, No.61976210, No.62076235, No.61806200, No.62002356, No.62002357, 
No.62006230.
\end{acks}

\bibliographystyle{ACM-Reference-Format}
\bibliography{sample-base}

\end{document}